# Ordinal-ResLogit: Interpretable Deep Residual Neural Networks for Ordered Choices

Kimia Kamal[1] and Bilal Farooq[1]

[1]Laboratory of Innovations in Transportation (LiTrans),
Ryerson University, Toronto, Canada

**Abstract**

This study presents an Ordinal version of Residual Logit (Ordinal-ResLogit) model to investigate the ordinal responses. We integrate the standard ResLogit model into COnsistent RAnk Logits (CORAL) framework, classified as a binary classification algorithm, to develop a fully interpretable deep learning-based ordinal regression model. As the formulation of the Ordinal-ResLogit model enjoys the Residual Neural Networks concept, our proposed model addresses the main constraint of machine learning algorithms, known as black-box. Moreover, the Ordinal-ResLogit model, as a binary classification framework for ordinal data, guarantees consistency among binary classifiers. We showed that the resulting formulation is able to capture underlying unobserved heterogeneity from the data as well as being an interpretable deep learning-based model. Formulations for market share, substitution patterns, and elasticities are derived. We compare the performance of the Ordinal-ResLogit model with an Ordered Logit Model using a stated preference (SP) dataset on pedestrian wait time and a revealed preference (RP) dataset on travel distance. Our results show that Ordinal-ResLogit outperforms the traditional ordinal regression model for both datasets. Furthermore, the results obtained from the Ordinal-ResLogit RP model show that travel attributes such as driving and transit cost have significant effects on choosing the location of non-mandatory trips. In terms of the Ordinal-ResLogit SP model, our results highlight that the road-related variables and traffic condition are contributing factors in the prediction of pedestrian waiting time such that the mixed traffic condition significantly increases the probability of choosing longer waiting times.

Email address: kimia.kamal@ryerson.ca, bilal.farooq@ryerson.ca

1. **Introduction**

Discrete Choice Models (DCMs) have been widely used to analyze decision makers' choices in various transportation areas such as mode choice, telecommuting frequency, number of non-works stops, travel distance, travel route, and car ownership. In many real-world classification problems, there exists an ordinal relationship between discrete alternatives of a choice set. In other words, each response has an inherent ordinal correlation with other alternatives. Akin to general classification problems, ordinal classifiers describe decision makers' preference towards alternatives over a discrete choice set but on an ordinal scale. In several transportation problems such as level of pedestrians' wait time, travel distance, and level of traffic accidents severity, individuals' responses follow a natural ordering. Therefore, modeling the ordinal form of alternatives is more representative of the real-world, understandable and describable compared to the continuous form (Kamal & Farooq, 2022; Michalaki et al., 2015; Parsa et al., 2019). To shed further light on the matter, the difference between, for instance, 1 and 1.7 seconds, may not be clear for researchers and modellers. In addition, the decision-makers usually categorize such inherently continuous variables into discrete categories and then make their decision. Hence the difference between low, medium and high categories broadly makes a better sense to researchers and stakeholders. Indeed, by discretizing a continuous dependent variable into different categories and more importantly, considering ordinal information among different categories, the effect of contributing factors on studied problems can better be evaluated.

Despite having a significant impact of ordinal information on the prediction task and model analysis, several studies simply ignored the ordinal form of responses in the classification (Mannering & Mokhtarian, 1995; Shankar & Mannering, 1996). They even applied behavioural models structured based on independently and identically distributed (IID) assumptions, resulting in the independence of irrelevant alternatives (IIA) property. This property is the most important feature of the Multinomial Logit (MNL) model, which may mistakenly be considered as an appropriate model structure for modelling discrete ordinal data due to interpretability and readily computational tractability. Nevertheless, we need an ordered-based discrete choice model for modelling ordinal information of ranking data. The traditional Ordered Logit model or Proportional Odds Model (POM) model is the most popular ordered discrete choice model (McCullagh, 1980). This model specification contemplates the natural ordering of responses and similar to the MNL model, not only does the ordered logit model enjoy simplicity in computation but researchers are able to describe and interpret the effect of contributing variables.

Since data-driven approach has recently led to considerably promising results in modelling, in this study, we set ourselves a goal to develop a deep learning-based ordered logit model, so as to overcome the main limitations of the traditional ordered logit model. Broadly, the data-driven approach enables modellers to capture the relationship between input and output directly from the data. Deep Neural Networks (DNNs), as a data-driven approach consisting of several layers, mimic the network structure of neurons in the human brain. These algorithms chiefly provide the feasibility and flexibility in terms of developing complicated non-linear models (Friston & Stephan, 2007). Moreover, the significant similarities between choice behaviour theory and DNN algorithms have led to great stride in the performance of discrete choice models (Cantarella & de Luca, 2005; Wong & Farooq, 2021). Despite achieving highly accurate prediction tools using neural networks, these models typically lack model interpretability, statistical analysis and behavioural indicators or economic information analysis (Karlaftis & Vlahogianni, 2011). However, in recent studies, not only have researchers tried to address these issues by combining DCMs with DNN frameworks (Wang, Mo, et al., 2020; Wong & Farooq, 2021), but also they have highlighted that these two modelling approaches are complementary in terms of prediction and interpretation (Wang, Wang, et al., 2020). To tackle the restrictions of neural networks, particularly in terms of interpretability and estimate far better accurate DCMs, recent research proposed a novel deep learning version of the multinomial logit model (MNL). Wong and Farooq (Wong & Farooq, 2021) employed Residual Neural Networks or ResNets architecture to capture underlying unobserved heterogeneity from the data. ResNets is a deep learning structure in which residual layers of the deep network, representing influential unobserved behaviour, are estimated independently by means of skipped connection (He et al., 2016).

Ordered transportation choices have rarely been modelled using these emerging data-driven techniques. Our study presents a logical continuation of ResLogit, designed explicitly for ordinal data. We define this model structure as an Ordinal-ResLogit model. Specifically, the Ordinal-ResLogit model combines ResLogit with a recent concept of COnsistent RAnk Logits (CORAL) (Cao et al., 2020). The CORAL framework is a consistent rank logits algorithm for the ordered nature of ranking data proposed by Cao et al. (2020). Hence, the Ordinal-ResLogit model benefits from three significant merits. First, similar to the ResLogit, our proposed model significantly represents real-world behaviours by considering unobserved heterogeneity in the choice selection process. Second, this model addresses the main limitation of machine learning algorithms known as black-box and provides the feasibility of interpreting explanatory variables for modelers. Last but not least, the resulting Ordinal-ResLogit structure is based on an extended binary classification approach—where, its theoretical formulation guarantees rank-monotonicity and consistency among binary classifiers.

To evaluate the performance of Ordinal-ResLogit model structure, we use both Stated Preference (SP) and Revealed Preference (RP) datasets. As the interaction between Automated Vehicles (AVs) and road users has been one of the most active and important areas of recent transportation research, we use a novel Virtual Reality (VR) dataset to predict the influence of AVs on pedestrian wait time in new traffic conditions, but with current road characteristics. VR is a developed SP survey in which computer-generated experiences are employed for simulation of futuristic scenarios (Farooq et al., 2018). The design of hypothetical scenarios in SP surveys, may put obstacles in the way of evaluation of model performance and accordingly leads to bias in results; therefore, we also analyze the ordinal discrete form of travel distance by categorizing RP travel distance data into five ordinal categories. RP datasets include real travel behaviour information limited to current circumstances and thus provide more reliable conclusions about the performance of our proposed model.

Generally, enhancing travel behavioural modelling is our major focus in this study and our main contributions to the transportation literature are:

- Formulation of Ordinal-ResLogit model and the associated indicators for economic analysis, which is classified as interpretable deep learning-based model for ordered choices
- Comparison of the performance of Ordinal-ResLogit model with the traditional ordered logit model to make a better conclusion for future research
- Evaluation of the performance of our proposed model by using both Stated Preference (SP) and Revealed Preference (RP) data.

The rest of this paper is organized as follows. First, a brief review of the literature on ordinal regression models and machine learning-based ordinal models are provided. It is followed by the methodology section in which the structure of Reslogit, CORAL and the Ordinal-ResLogit model is described in detail. In section 3, a description of RP and SP data is presented and then analysis of estimated models is provided in section 4. Finally, in the conclusion section, some suggestions and future directions of the study are proposed.

## 2. Background

In this section, we first present a general overview on conventional ordinal regression models. Secondly, since the main focus of this study is on deep learning-based ordinal choice models, a review of ordinal regression model development using machine learning algorithms is provided, in the following sections.

### 2.1. Ordinal Regression Models

Ordinal regression models are widely used to solve classification problems in which the labels of categories provide enough information to order. Using an ordinal classification method for modeling ordinal data dates back to 1980, where McCullagh proposed multivariate extensions of generalized linear models, including Proportional Odds Models (POM) or Ordered Logit Models (McCullagh, 1980). This model is the most popular approach for modeling the ordered nature of ranking data in which an unobserved continuous variable, known as a latent variable, relates to ordinal responses through thresholds or cut points. In other words, the ordered logit model is an ordinal regression model for a discrete dependent variable existing an ordinal relationship among alternatives. The basic structure of proportional odds models or threshold models for a set of ordinal responses, $r_1 < r_2 < \cdots < r_k$ can be mathematically represented as follows:

$$U_n^* = \beta x_n + \eta_n \tag{1}$$

$$U_n = r_k \ if \ \delta_{k-1} < U_n^* \leq \delta_k \qquad for \ k = 1, 2, \ldots. K$$

Where $U_n$ represents the ordinal response and $U_n^*$ is the latent variable which is assumed to be a linear function of explanatory variables ($x_n$) with associated parameters ($\beta$), which remain constant between ordinal categories. It is of note that the non-linear function can also be applied; however, due to simplicity, the linear form is a dominant form of random utility function in discrete choice analysis. In equation (1), $\eta_n$ is the random error component of the utility function such that the logistic distribution is the most common choice for $\eta_n$, resulting in an ordered logit model. Regarding the mentioned formulation, the ordinal regression problem can be considered between regression and multi-class classification, but there are two main differences between them and the ordinal regression model (Gutiérrez et al., 2015). First, in contrast to classification problems, there is an ordinal relationship between different categories. Second, unlike the regression problems, distances between categories are defined arbitrarily.

This model is based on the cumulative probabilities of the response variable, where the cumulative probabilities are related to a latent variable through the logit function. The cumulative choice probability can be written as:

$$P(U_n^* > \delta_k) = \frac{1}{1 + \exp(\beta x_n - \delta_k)} \qquad (2)$$

And the choice probability of choosing $k^{th}$ alternative in the set of ordinal responses ($r_1 < r_2 < \cdots < r_k$) for individual $n$ is simply obtained as follows:

$$P(U_n = r_k) = P(U_n^* > \delta_k) - P(U_n^* > \delta_{k-1}) \qquad (3)$$

$$= \frac{1}{1 + \exp(\beta x_n - \delta_k)} - \frac{1}{1 + \exp(\beta x_n - \delta_{k-1})}$$

In this model specification, a set of thresholds ($\delta_k$) plays an important role for classification and associates the latent variable ($U_n^*$) to ordinal response ($U_n$). It is of note that thresholds must satisfy the constraint $\delta_1 \leq \delta_2 \leq \cdots \leq \delta_k$ in which $k$ is the index of each label. In other words, label $r_k$ is observed if and only if $U_n^* \in [\delta_{k-1}, \delta_k]$. As a result, two groups of parameters are estimated through this methodology from the data: the vector of parameters are associated with explanatory variables ($\beta$) and a set of thresholds ($\delta_k$). It is worth mentioning that adding an arbitrary constant to the latent variable can be counteracted by subtracting the same constant from each threshold. This identification problem can be solved by either removing the constant from the latent variable or fixing the first threshold to zero.

The constant effect of variables on ordinal categories is known as the parallel regression assumption of the ordered logit model, which is the most popular model framework for ordinal data in the literature. This limitation leads to equal odds ratios between each pair of choices across ordinal categories. Due to this assumption, the ordered logit model is also known as Proportional Odds Models (POM). This assumption can be relaxed by allowing explanatory variables to have alternative-specific parameters in the function of latent variable (Eluru et al., 2008; Yasmin et al., 2014). In addition, the latent variable can be the combination of alternative-specific and generic parameters. This flexible model specification is known as the Generalized Ordered Logit (GOL) model in the literature. Ordered Generalized Extreme Value (OGEV) is an ordered version of GEV model in which alternatives are allocated to nests based on their ordering relationship (Small, 1987). It is of note that in contrast to the traditional ordered logit model and its variants, OGEV is free from the estimation of thresholds and this issue makes this model more flexible.

Although GOL and OGEV models structures can address the main limitation of the ordered logit model, the full unobserved heterogeneity among ordinal categories is not captured through these frameworks. Therefore, since the advent of deep neural network algorithms, modellers strive to optimize the performance of discrete choice models by capturing unobserved heterogeneity and minimizing the difference between decision-makers' behaviour and model prediction.

## 2.2. Ordinal Regression Machine Learning

Several fields of science and engineering have achieved significant improvement in estimation and evaluation by the promising performances of machine learning algorithms. The main purpose of machine learning algorithms is to relax some strict assumptions in traditional models and improve the accuracy of prediction (Vythoulkas & Koutsopoulos, 2003). The general structure of the utility function of traditional DCMs is based on two key components: 1) an observed deterministic component representing the effect of explanatory variables and 2) an observed error term capturing the unobserved factors (Ben-Akiva, M.E., Lerman, 1985). However, the utility specifications are defined in advance of estimating the model according to the behavioural decision theories or modellers' prior knowledge (Hillel et al., 2020). Although the vital importance of accounting uncertainty or error component in behavioural modeling is highlighted in studies (Sims, 2003), the fixed assumptions for the distribution of errors prevent the model structure from fully capturing information from the data and thus lead to ignoring the effect of heterogeneity in the choice modeling data. In other words, in conventional DCMs, on the one hand, the actual decision-makers' behaviour is not directly observed, and on the other hand, a pre-specified model structure gets in the way of considering total underlying unobserved factors from the data. Hence, in recent years, the rapid emergence of data-driven approaches has provided an opportunity for researchers to achieve a more flexible model structure and further enhance the performance of discrete choice modeling.

In this regard, the focus of some recent studies has been to propose machine learning-based ordinal regression models in order to achieve more powerful tools for the prediction of ordinal data. Different machine learning approaches such as support vector machine learning (Chu & Keerthi, 2005; Herbrich et al., 1999) and perceptron learning algorithm (Shen & Joshi, 2005) have been used for a modification of ordinal regression models in the literature. Some research strictly focused on the structure of threshold models and some others tried to use classification algorithms due to their simplicity in implementation. Li and Lin proposed a reduction framework in which the ordinal regression problem is transformed into multiple binary classification sub-problems (Lin & Li, 2007). The reduction framework unifies the idea of some exiting ordinal regression approaches. However, this framework depends on a misclassification cost matrix in which three issues should be taken into account. Let $C_{y,r_k}$ be an array of cost matrix $C$, measuring the cost of predicting a training example $(x, y)$ as $r_k$. First, the function of the cost matrix must be defined based on prior knowledge of modelers. For instance, in ordinal regression models, the difference between actual and predicted categories is often considered as a cost matrix. Second, the cost of each binary task is calculated separately for each individual, causing a computationally expensive training process. More importantly, to achieve a rank-monotonic threshold model, a cost matrix must be convex. It means that the framework should satisfy the basic requirement in which $C_{y,r_{k-1}} \geq C_{y,r_k}$ if $r_k \leq y$ and $C_{y,r_k} \leq C_{y,r_{k+1}}$ if $r_k \geq y$.

In the field of computer science, in order to predict human age, theoretical algorithms have been employed to develop DNN versions of ordinal regression models. Recent research proposed a learning-based ordinal regression algorithm called OR-CNN by applying the reduction framework in Convolutional Neural Network (CNN) (Niu et al., 2016). This method is an end-to-end modeling approach and can achieve state-of-the-art results. Nonetheless, OR-CNN suffers from inconsistency among binary classifiers tasks (Niu et al., 2016), meaning that based on the ordinal relationship, the output of the $k^{th}$ binary task may belong to lower categories in comparison to the prediction of its previous task. To address this problem, Chen et al. developed an ensemble-based algorithm called Ranking-CN (Chen et al., 2017). Ranking-CN is an ensemble of CNNs for binary classification and aggregated their binary output to predict the age label. They showed that Ranking-CNN outperforms individual binary classifiers. COnsistent RAnk Logits (CORAL) framework is another deep learning ordinal regression approach that is primarily based on the structure of both threshold models and binary classification algorithms (Cao et al., 2020). The CORAL structure is classified as a rank-monotonic threshold model, whereas it is free from constraints of the reduction framework. In fact, the CORAL framework does not require a cost matrix conditioned by convex conditions and depends on a priori knowledge. In addition, this framework can readily be implemented in various neural network architectures.

Our Ordinal-ResLogit approach combines the CORAL framework with the ResLogit, so as to propose a deep learning version of the threshold model. ResLogit is primarily a learning-based multinomial logit model that enjoys ResNet architecture, leading to easier optimization and further accuracy utilizing increased depth (Wong & Farooq, 2021).

## 3. Methodology

In the following section, first, the structure of ResLogit and CORAL algorithms is briefly described and then our proposed model is presented.

### 3.1. ResLogit Model Structure

The ResLogit model is a novel learning behaviour algorithm for modeling multinational logit problems. The utility specification of this model has a correction for heterogeneity in the error component to achieve a more robust utility function, representing real-world human behaviour (Wong & Farooq, 2021). Broadly, the Reslogit model chiefly focused on the independence of irrelevant alternatives (IIA) assumption which has been the primary property of discrete choice modelling. To correct this assumption, Wong and Farooq applied Residual Neural Networks architecture in which the deep neural network is applied into residual layers, instead of the whole structure of the model. As a result, not only can Reslogit model capture random heterogeneity from the data, but also the deterministic term of the utility function of traditional discrete choice models is retained (Wong & Farooq, 2021). Therefore, in the ResLogit model, the utility of the individual $n$ choosing choice $k$ from $K$ alternatives is defined by:

$$U_{kn} = V_{kn} + g_{kn} + \varepsilon_{kn} \tag{4}$$

where $V_{kn}$ is a deterministic component consisting of explanatory variables, $g_{kn}$ represents residual component capturing random heterogeneity and $\varepsilon_{kn}$ is the error term capturing remaining unobserved errors. A striking characteristic of ResLogit is that it allows us to evaluate the effect of any changes in parameters associated with independent explanatory variables, due to the structure of residual layers using skipped connection. Therefore, this learning model is classified as an interpretable deep learning model. Furthermore, The ResLogit model does not suffer from the vanishing gradient problem, due to the skip structure, allowing each residual layer to be estimated independently (Wong & Farooq, 2021).

In ResLogit, the residual layers, capturing heterogeneity, are associated with the entropy concept such that a state with low entropy is an optimal state (Wong & Farooq, 2021). Therefore, the equation (4) can be rewritten for $m$ residual layers (Wong & Farooq, 2021):

$$U_{kn} = V_{kn} - \sum_{m=1}^{M} \ln\left(1 + \exp(W^m \cdot V_n^{(m-1)})\right) + \varepsilon_{kn} \qquad for \ m = 1, 2, \dots M \tag{5}$$

In equation (5), $V_n^{(m-1)}$ is the output vector of non-linear utility components for the $m-1^{th}$ residual layer consisting of $K$ members and is formulated as:

$$V_n^m = V_n^{m-1} - \ln\left(1 + \exp(W^m \cdot V_n^{(m-1)})\right) \qquad for \ m = 1, 2, \dots M \tag{6}$$

Our study assumes that residual connection comes from the previous layer. In addition, $V_n^{(0)}$ is the vector of deterministic component of alternatives or linear function of observed variable ($V_n$), and $W^m$ is a matrix $K \times K$ of residual parameters representing heterogeneity captured by $m^{th}$ residual layer.

3.2. COnsistent RAnk Logits (CORAL) Framework

The CORAL framework can be classified as a learning-based ordinal regression model that inherits the theoretical concept of both threshold models and binary classification algorithms (Cao et al., 2020). It is a single deep neural network such that its output layer consists of $K - 1$ binary classifiers. To reduce the ordinal problem to multi binary classification, for each $(x_n, y_n)$ training example, first, a rank $y_n$ is extended into $K - 1$ binary labels $y_n^1, y_n^2, \dots, y_n^{k-1}$ using a direct question "is $y_n$ greater than $r_k$?" Considering indicator index $I$ that can assume values of 0 and 1, $y_n^k$ can be defined by equation (7). $I$ equals 1 when the expression inside it is true and otherwise it is 0.

$$y_n^k = I[y_n > r_k] \tag{7}$$

The main purpose of CORAL structure is to guarantee consistency among multiple binary classifiers using the same weight parameters ($W_k$), for $K - 1$ binary classifiers (Cao et al., 2020). Let σ(x) be the logistic sigmoid function as follows:

$$\sigma(x) = \frac{1}{1 + \exp(-x)} \tag{8}$$

After adding independent bias parameters to the input of the final layer ($b_k$), each binary classifier is able to predict the probability of $y_n > r_k$ or $y_n^k = 1$ based on logistic sigmoid function:

$$P(y_n^k = 1) = \sigma\left(\left(\sum_{k=1}^{K} W_k U_k\right) + b_k\right) \tag{9}$$

Where $U_k$ is the utility of alternative $k$ which is obtained from the penultimate layer of the neural network framework. Non-increasing bias parameters and consequently non-increasing predicted probabilities ($P(y_n^1) > P(y_n^2) ... > P(y_n^{K-1})$) ensure the consistency between binary classifiers (Cao et al., 2020).

### 3.3. Ordinal-ResLogit Model

We propose a method of combining ResLogit with the CORAL framework. We define the proposed model as the Ordinal-ResLogit model. In the Ordinal-ResLogit model, the utility function of ordinal categories for individual $n$, representing latent variable in the ordinal regression model, is obtained from equation (5) in which the heterogeneity component is captured by means of residual layer. However, the final layer of our proposed model is fed into the CORAL framework. The Ordinal-ResLogit model structure is illustrated in Figure 1.

The Ordinal-ResLogit model is also based on the cumulative probabilities of the response variable and the choice probability of choosing $k^{th}$ alternative is:

$$P(y_n = r_k) = P(y_n > r_k) - P(y_n > r_{k-1}) \tag{10}$$

$$= P(y_n^k = 1) - P(y_n^{k-1} = 1)$$

To model a discrete choice problem, a choice set needs to be exhaustive, meaning that each individual must necessarily choose one of the alternatives. Therefore, In Ordinal-ResLogit model, we suppose that the first category is always chosen, and eventually, based on the prediction of $K-1$ binary classifier task, the rank index of each individual is predicted as follows:

$$\hat{y}_n = 1 + \sum_{k=1}^{K-1} f_k(x_n) \tag{11}$$

Where $f_k(x_n) \in \{0,1\}$ is the prediction of $k^{th}$ binary classifier such that $f_k(x_n) = 1$ if $P(y_n^k = 1) > \alpha$. The value of threshold $\alpha$, ranging from 0 to 1, determines the prediction of binary classifiers, and in binary classifications, this threshold is usually considered equal to 0.5. However, in this study, we try to choose the best value of $\alpha$ parameter according to the accuracy of our proposed model.

Regarding the structure of Ordinal-ResLogit model, this model provides an opportunity to capture heterogeneity from the data and exploit the interpretability of the ResLogit structure, which is rooted in the operation of residual layers. In fact, even if the effect of explanatory variables is constant among ordinal alternatives, the residual matrices consider the heterogeneity and correlation between the categories, resulting in a specific-utility function for each alternative. Hence, Ordinal-ResLogit allows us to overcome the main limitation of ordered logit model rooted in parallel regression assumption. In addition, the Ordinal-ResLogit model offers a rank-monotonic threshold model for the evaluation of ordinal data using the CORAL framework. It is noticeable that although the CORAL framework is designed for ordinal data using a binary classification method, its structure does not require a misclassification cost matrix. In ordinal problems, it is self-evident that the cost values are not the same for different prediction errors. For instance, if the actual choice of a pedestrian is low, the error cost of predicating high should obviously be more than medium. As a result, the formulation of the cost matrix strongly influences the performance of the model.

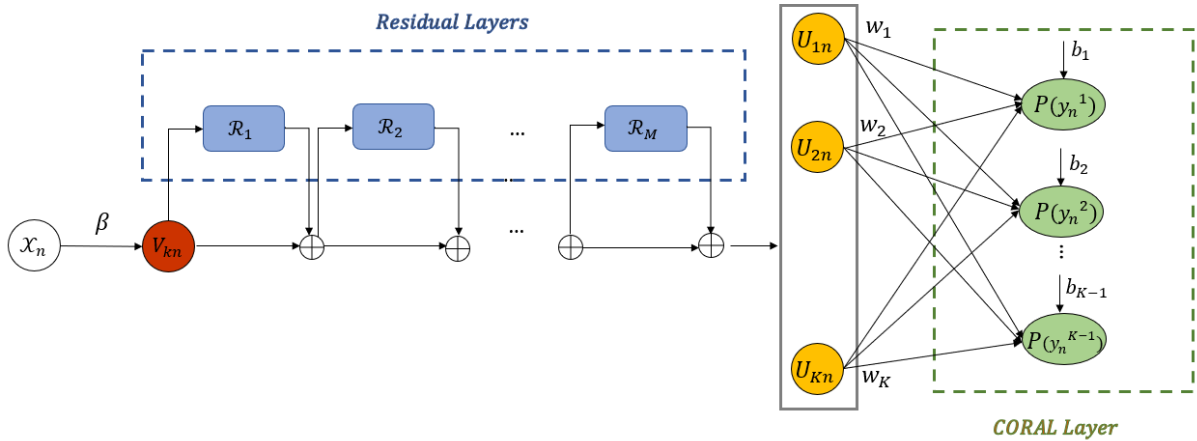

Figure 1: The architecture of Ordinal-ResLogit model

### 3.3.1. Model Estimation

To estimate the set of optimal parameters in the Ordinal-ResLogit model, 1) parameters associated with explanatory variables ($\beta$), 2) weights of residual layers and CORAL layer (final layer) ($W^m, W_k$) and 3) bias parameters ($b$), the loss function shown in equation (10) should be minimized:

$$LL(\beta, W, b) = -\sum_{n=1}^{N} \sum_{k=1}^{K-1} \lambda^{(k)} [\ln(P(y_n^k = 1))y_n^k + \ln(1 - P(y_n^k = 1))(1 - y_n^k)] \quad (12)$$

$\lambda^{(k)}$ denotes the classifier task importance weight in estimation. For simplicity, we assume for all tasks $\lambda^{(k)} = 1$ and as a result $K - 1$ binary classifiers have the same importance in optimization. However, the only requirement for guaranteeing rank monotonicity is that $\lambda^{(k)}$ is positive (Cao et al., 2020).

We used a data-driven stochastic gradient descent-based (SGD) learning algorithm and applied an RMSprop optimization step to scale the learning rate based on model parameters, since this hyperparameter has a significant impact on model performance. In the data-driven optimisation algorithm, 0-1 validation error is reduced by indirectly minimizing the loss function of the training dataset. The data-driven optimization method is an efficient estimation method for machine learning-based models with numerous estimated parameters (Goodfellow et al., 2016). The structure of skip connection prevents the vanishing gradient problem during computing the derivative of the loss function with respect to parameters by using backpropagation (Wong & Farooq, 2021). Therefore, the gradient of residual layers is independent of the derivative of deterministic component and parameters associated with explanatory variables can always be updated throughout the estimation.

### 3.3.2. Model Evaluation

For model evaluation and comparison, we use Mean Prediction Error (MPE) on the validation dataset as follows:

$$MPE = \frac{1}{N_{validation}} \sum_{n=1}^{N} I[y_n \neq \hat{y}_n] \quad (13)$$

where $y_n$ and $\hat{y}_n$ are the actual choice and predicted rank respectively and indicator index $I$ equals 1 when the expression inside it is true and otherwise, it is 0. In addition to prediction accuracy, the performance of the Ordinal-ResLogit model and traditional ordered logit model can be compared based on the Akaike Information Criterion (AIC) which is a common measure for comparing different models which are different in the number of estimated parameters. In modelling process, there is a trade-off between goodness of fit and simplicity of a model which is rooted in the number of estimated parameters. AIC evaluates the performance of models based on these two factors and is calculated by:

$$AIC = -2LL(\beta, W, b) + 2B \tag{14}$$

Where, $LL(\beta, W, b)$ is the maximum log-likelihood function, representing goodness of fit and B is the number of estimated parameters. In general, a model with a lower AIC shows relatively better performance.

## 4. Case Study

### 4.1. Data

To provide better evaluation about performance of Ordinal-ResLogit, we use two types of datasets, 1) Stated Preference (SP) on wait time from virtual reality and 2) Revealed Preference (RP) on travel distance from London, and compare their results with the conventional ordered logit model.

First, a virtual reality dataset, which is an advanced SP data and collected based on the Virtual Immersive Reality Experiment (VIRE), is employed. VIRE is a virtual reality simulation framework aimed to immerse the participants in an artificial 3D environment where participants are far immersed in experiments as they feel in the real world (Farooq et al., 2018). Particularly, VIRE technology provides an opportunity for researchers to analyze futuristic scenarios in which pedestrians will be exposed to AVs. VIRE allows us to seek the effect of AVs on pedestrian behaviour, and also this technology keeps pedestrians safe in the experiments. Kalatian and Farooq designed 90 scenarios defined by 9 controlled variables categories: rules and regulations, street characteristics, automated vehicles feature, traffic demand and environmental conditions (Kalatian & Farooq, 2021). Detailed information about the definition of scenarios can be found in their study. However, two important points should be taken into account while using this data: (1) In the experiments, participants can distinguish human-driven from automated vehicles and (2) Although the difference between the braking system of AVs and human-driven vehicles was considered in the experiments, participants might not be realized, unless during the experiment (Kalatian & Farooq, 2021). The data collection process was conducted over 6 months, from April 2018 to September 2018. During this period, 113 adults and adolescents, and 47 kids and teenagers participate in experiments in which 15 scenarios were randomly chosen for each participant. This data comprises crossing information related to street characteristics, traffic condition and environment situation as well as socio-demographic and travel patterns, gathered through questionnaires. Overall, after pre-processing the data and eliminating child responses, there are 2,291 responses for the prediction of pedestrian waiting time. A list of explanatory variables which are used for developing the Ordinal-ResLogit model is presented in Table 1.

As the main purpose of this study is discrete choice analysis of ordinal categories of pedestrian wait time and travel distance, we first need to discretize the continuous data into ordinal groups. To optimize the thresholds of different categories for dependent variables, the Jenks Natural Breaks classification method is applied (Jiang, 2013). This method classifies a given data into several groups in such a way as to minimize the variance of members of each group while maximizing the variance between different groups.

By applying the Jenks Natural Breaks method to the wait time variable, only 1.86% of participants in virtual reality experiments waited more than 33.00 seconds which is the threshold of the high category. To obtain more balanced data and realistic representation, we try to optimize the given thresholds based on logical facts. In the real world, the minimum waiting time is assumed to be the sum of perception and reaction time. The perception time is approximately 1.5 to 3 seconds for pedestrians and often after additional time, around 2 seconds, they start physically crossing. In addition, based on the fact that crossing two lanes usually takes between 20-22 seconds, the threshold of high waiting time can be rationally optimized. Finally, the given pedestrian wait time data is classified into three discrete groups: low: pedestrians who wait less than 5 seconds, medium: a group of pedestrians who has waiting time between 5 to 20 seconds on the sidewalk, and high: pedestrians waiting more than 20 seconds. The frequencies of crossing with different waiting time are shown in Figure 2.

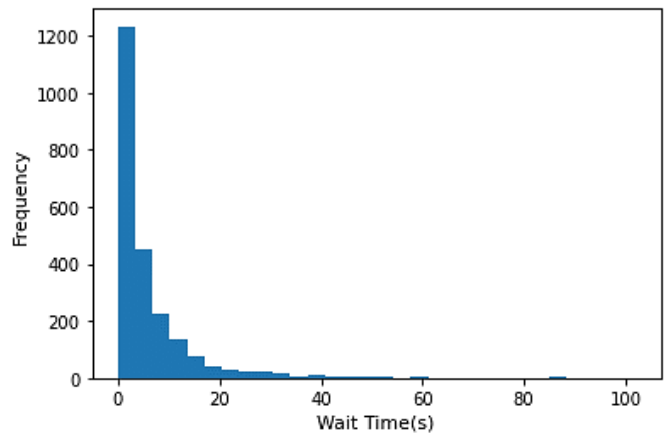

Figure 2: Wait time frequency

Furthermore, we evaluate the performance of Ordinal-ResLogit structure by analyzing ordinal categories of travel distance which can provide better insight into the impact on decision makers' travel behaviour by using a RP travel dataset. We use an available open RP dataset of the Travel Demand of the metropolitan of London. This data was developed by combining individual trip records of the London Travel Demand Survey (LTDS), from April 2012 to March 2015, with systematically matched trip trajectories alongside their corresponding mode alternatives from a directions Application Programming Interface (API) (Hillel et al., 2018). Non-mandatory trips are selected for this study since the destination is not fixed for this type of trips and decision-makers have enough freedom to choose among different distances according to their preference and trip attributes. The description of explanatory variables used in this study is shown in Table 2. The data comprises socio-demographic and travel attributes extracted from LTDS survey and the API directions service.

Table 1: Description of explanatory variables of SP pedestrian wait time data

| Variable | Description | Mean | Standard deviation |
|---|---|---|---|
| **_Traffic condition_** | | | |
| Mixed traffic condition | 1: if traffic in scenario consists of automated vehicles and human-driven vehicles, 0: otherwise | 0.041 | 0.199 |
| Fully automated condition | 1: if traffic in scenario consists of only automated vehicles, 0: otherwise | 0.933 | 0.25 |
| Human-driven condition | 1: if traffic in scenario consists of only human-driven vehicles, 0: otherwise | 0.025 | 0.157 |
| **_Street characteristics_** | | | |
| Low lane width | 1: if the lane width is less than 2.75 meter, 0: otherwise | 0.344 | 0.475 |
| High lane width | 1: if the lane width is greater than 2.75 meter, 0:otherwise | 0.349 | 0.477 |
| Two way with a median | 1: if the road type is two way with median, 0: otherwise | 0.316 | 0.465 |
| Two way | 1: if the road type is two way, 0: otherwise | 0.342 | 0.475 |
| One way | 1: if the road type is one way, 0: otherwise | 0.341 | 0.474 |
| Density | Density of road (veh/hr/ln) | 20.474 | 7.304 |
| **_Socio-demographic_** | | | |
| Age 18-30 | 1: if participant's age is between 18 and 30, 0: otherwise | 0.477 | 0.49 |
| Age 30-39 | 1: if participant's age is between 30 and 39, 0: otherwise | 0.361 | 0.481 |
| Age 40-49 | 1: if participant's age is between 40 and 49, 0: otherwise | 0.079 | 0.269 |
| Age over 50 | 1: if participant's age is more than 50, 0: otherwise | 0.083 | 0.276 |
| Female | 1: if participant is female, 0: otherwise | 0.432 | 0.495 |
| Driving license | 1: if participant has a driving license, 0: otherwise | 0.916 | 0.278 |
| No car | 1: if participant has no car in the household, 0: otherwise | 0.219 | 0.414 |
| One car | 1: if participant has one car in the household, 0: otherwise | 0.36 | 0.48 |
| Over one car | 1: if participant has more than one car in the household, 0: otherwise | 0.421 | 0.494 |
| Active mode | 1: if participant uses active modes regularly, 0: otherwise | 0.233 | 0.423 |
| Private car mode | 1: if participant uses private car regularly, 0: otherwise | 0.337 | 0.473 |
| Public mode | 1: if participant uses transit regularly, 0: otherwise | 0.43 | 0.495 |
| Walk to work | 1: if participant usually walks to work, 0: otherwise | 0.409 | 0.492 |
| Walk to shop | 1: if participant usually walks to shop, 0: otherwise | 0.722 | 0.448 |
| VR experience | 1: if participant has VR experience, 0: otherwise | 0.419 | 0.494 |
| **_Environment condition_** | | | |
| Night | 1: if the time of scenario is night, 0: otherwise | 0.341 | 0.474 |
| Snowy | 1: if the weather of scenario is snowy, 0: otherwise | 0.316 | 0.465 |

Table 2: Description of explanatory variables of RP travel distance data

| Variable | Description | Mean | Standard deviation |
|---|---|---|---|
| *Travel attributes* | | | |
| Drive | 1: if individual uses private car, 0: otherwise | 0.478 | 0.450 |
| Public transportation | 1: if individual uses public transportation, 0: otherwise | 0.296 | 0.456 |
| Cycle | 1: if individual cycles, 0: otherwise | 0.023 | 0.151 |
| Walk | 1: if individual walks, 0: otherwise | 0.202 | 0.402 |
| Transit cost | Travel cost for transit mode (£) | 1.456 | 1.287 |
| Driving cost | Total travel cost for private car mode (£) | 1.423 | 2.853 |
| *Socio-demographic* | | | |
| Age 18-30 | 1: if individual's age is between 18 and 30, 0: otherwise | 0.185 | 0.388 |
| Age 30-45 | 1: if individual's age is between 30 and 45, 0: otherwise | 0.315 | 0.465 |
| Age 45-60 | 1: if individual's age is between 45 and 60, 0: otherwise | 0.232 | 0.422 |
| Age over 60 | 1: if individual's age is more than 60, 0: otherwise | 0.268 | 0.443 |
| Female | 1: if individual is female, 0: otherwise | 0.557 | 0.497 |
| Disb | 1: if traveler has disability; 0: otherwise | 0.233 | 0.423 |
| Driving license | 1: if individual has a driving license, 0: otherwise | 0.718 | 0.450 |
| No car access | 1: if there is no car in the household, 0: otherwise | 0.302 | 0.459 |
| Restricted car access | 1: if there is less than one car per adult in the household, 0: otherwise | 0.428 | 0.495 |
| Unrestricted car access | 1: if there is one or more than one car in the household, 0: otherwise | 0.271 | 0.444 |

Regarding the RP data, we also categorize the continuous travel distance variable into five ordinal groups of very short, short, medium, long and very long. The optimum thresholds and percentages of each category are shown in Table 3.

Table 3: Ordinal travel distance categories

| Category | Optimum thresholds (Kilometer) | Percentage |
|---|---|---|
| Very short | 0 - 7.8 | 47.1% |
| Short | 7.8 - 15.3 | 28.7% |
| Medium | 15.3 - 26 | 14.3% |
| Long | 26 - 41.4 | 7.1% |
| Very long | 41.4 – 94.7 | 2.9% |

*4.2. Model Hyperparameters*

In our Ordinal-ResLogit model, we divided the dataset into two subsets using a 70:30 training/validation split. Regarding the depth of the model, although there is an argument about the effect of hidden layers in the performance of deep learning networks and estimation process, increasing the number of residual layers in the Ordinal-ResLogit does not lead to degradation accuracy problem or overfitting due to the ResNet architecture (He et al., 2016; Wong & Farooq, 2021). After testing the performance of our model in terms of prediction accuracy with $M = \{2, 4, 8, 16\}$ and considering training computation time, we choose 16 residual layers. A minibatch stochastic gradient descent (SGD) learning algorithm with a batch size of 64 is used to train the models. As regularization techniques may improve the performance of learning models and we intend to compare our proposed model with the traditional ordered logit model and, we did not implement any forms of regularization, except the early-stopping approach in which the training process stops once the performance of the model on the validation dataset starts to degrade. Another effective hyperparameter in our proposed model is the threshold $\alpha$, determined the rank index of each individual. In fact, the prediction of each binary classifier relies on the value of this threshold. After examining the evaluation measure for different threshold values, ranging from 0.3 to 0.6, we ultimately chose 0.4 and 0.5 for SP and RP data respectively.

## 5. Result

*5.1. Model Performance*

The results obtained from two estimated models, the Ordinal-ResLogit and the traditional ordered logit models, for ordinal alternatives of pedestrian wait time and travel distance are presented in Table 4 and Table 5 respectively. We chose the ordered logit model as the base model structure due to several reasons. First, technically, we wanted to compare the performance of our learning-based model in a robust manner rather than in specific conditions. For instance, comparison with GOL model may not have been suited as that would be limited to incorporating alternative specific parameters only. Secondly, employing the ordered logit model is still more popular than any other option for modelling ordinal data. As in this study, we try to propose an interpretable deep learning-based model for ordinal data; we compare and evaluate the performance of models based on two measures: 1) the goodness of fit and models accuracy 2) the impact of explanatory variables.

In the Ordinal-ResLogit model, residual layers allow us to contemplate the effect of underlying unobserved behaviour in the prediction. In other words, the weight parameters of residual layers represent unobserved correlations between ordinal categories. Thus in the Ordinal-ResLogit model, the number of estimated parameters is considerably higher than the ordered logit model. Notably, there is a major trade-off between capturing unobserved heterogeneity by the means of deep learning algorithms and complexity of models which is considered as a penalty in our model selection method, AIC.

Due to very large number of parameters and their adverse effect on the complexity of the model, the Ordinal-ResLogit model for pedestrian wait time obtained a slightly higher AIC value than the ordered logit model. Regarding explanatory variables, the results show that, the deterministic component is constant among low, medium and long waiting time categories. However, the residual term of the utility function captured unobserved variables and correlation patterns among ordinal alternatives. In the ordered logit model, the effects of some variables, such as the number of cars, walking habits, night condition, are estimated insignificant while the results confirm the highly significant impact of these variables in the Ordinal-ResLogit model. In terms of prediction accuracy, there is no considerable difference between accuracy of two estimated models. However, as the performance of deep learning-based model considerably depends on hyperparameters setting we observed that our proposed model improves the prediction accuracy by 2.2% in comparison to ordered logit model in higher batch size. Hence, the result of SP data demonstrates that our proposed model generally outperforms the standard ordered logit model, although modelers must be aware of the inevitable impact of hyperparameters on the performance of deep learning-based models. Broadly, the Ordinal-Reslogit model provides the feasibility of modeling real-world pedestrian behaviour as well as retaining the deterministic component of the utility function.

Table 4: Results of estimated models for SP pedestrian wait time data

| Variable | Ordinal-ResLogit | | Ordered Logit | |
|---|---|---|---|---|
| | Value | t-stats | Value | t-stats |
| *Traffic condition* | | | | |
| Mixed traffic condition | 0.157 | 1.151 | 0.548 | 1.135 |
| Fully automated condition | -0.048 | -3.058 | 0.598 | 1.410 |
| *Street characteristics* | | | | |
| Low lane width | -0.257 | -9.796 | -0.344 | -2.515 |
| High lane width | 0.170 | 6.188 | 0.164 | 1.196 |
| Two way with a median | -0.204 | -7.557 | -0.274 | -1.974 |
| One way | 0.231 | 8.401 | 0.201 | 1.542 |
| Density | 0.080 | 9.976 | 0.053 | 7.229 |
| *Socio-demographic* | | | | |
| Age 30-39 | -0.328 | -12.268 | -0.812 | -6.371 |
| Age 40-49 | - | - | -0.831 | -3.515 |
| Age over 50 | 0.359 | 5.768 | 0.348 | 1.629 |
| Female | 0.352 | 14.411 | 0.717 | 6.106 |
| Over one car | 0.227 | 9.318 | - | - |
| Walk to work | 0.235 | 8.996 | 0.191 | 1.603 |
| Active mode | 0.080 | 2.459 | - | - |
| Public mode | -0.242 | -9.734 | -0.719 | -6.151 |

| Variable | Ordinal-ResLogit | | Ordered Logit | |
|---|---|---|---|---|
| | Value | t-stats | Value | t-stats |
| *Environmental variables* | | | | |
| Night | -0.174 | -6.859 | - | - |
| Snowy | 0.121 | 4.195 | 0.122 | 1.029 |
| Threshold1 (Bias1) | 1.920 | - | 2.019 | 4.379 |
| Threshold2 (Bias2) | 0.246 | - | 4.396 | 9.423 |
| No. observation | 2,291 | | 2,291 | |
| No. parameters | 165 | | 16 | |
| Log-likelihood | 1276.57 | | 1193.68 | |
| AIC | 2883.16 | | 2419.36 | |
| Validation accuracy | 61.19% | | 61.92% | |

Regarding the results of RP data, the AIC value of our proposed model is significantly lower than the ordered logit model. In addition, there is a considerable difference between the accuracy of the two estimated models, approximately 26%. The result highlights that Ordinal-Reslogit allows modelers to achieve more accurate predictions by capturing the impact of unobserved variables that undoubtedly affect decision-maker behaviour in the real world. Interestingly, RP data, which are more reliable in choice analysis, help us to show the strong performance of Ordinal-ResLogit. It is of note that for five ordinal categories of travel distance, the deterministic components of utility function are also estimated with different values, meaning that the observed variables have a different effect on alternatives. Therefore, like the generalized ordered logit model, our proposed model allows explanatory variables to have alternative-specific parameters. Furthermore, in the Ordinal-Reslogit, the probability of each alternative is computed based on pairwise comparison and each alternative is compared with other choices; therefore, in our proposed model, the utilities of alternatives are in ordinal form. Similar to SP data, the higher significant impact of some variables is observed in the Ordinal-ResLogit model compared to the ordered logit model. It is worth mentioning in this data the t-statistic information of some variables is noticeably high. In other words, their standard error is very low. This result may be rooted in the training process since the performance of our proposed model significantly relies on hyperparameters. According to our evaluation, the value of batch size considerably affects t-statistics in such a way that a higher batch size results in higher values. However, in our study, by considering a lower batch size, the contributing factors remain at a significant level. In short, the results of RP travel distance further underline the benefits of our proposed model for modelling ordinal data.

Table 5: Results of estimated models for RP travel distance data

| Variable | Ordinal-ResLogit | | | | | Ordered logit |
|---|---|---|---|---|---|---|
| | Very short | Short | Medium | Long | Very long | |
| *Travel attributes* | | | | | | |
| Transit | 0.256 | -0.469 | -0.224 | 0.575 | 0.012 | 0.754 |
| | (89.625)* | (-115.556) | (-23.713) | (58.805) | (1.789) | (23.566) |
| Cycle | 0.463 | -0.281 | -0.522 | 0.134 | -0.213 | -0.398 |
| | (45.021) | (-20.504) | (-17.035) | (4.193) | (-9.461) | (-5.242) |
| Walk | 1.180 | -0.052 | -1.184 | -0.632 | -0.951 | -2.512 |
| | (206.162) | (-7.328) | (-97.058) | (-40.095) | (-92.547) | (-51.265) |
| Driving cost | 0.348 | 0.470 | 0.833 | 0.793 | 0.832 | 0.246 |
| | (502.319) | (411.501) | (311.589) | (243.783) | (365.519) | (61.450) |
| Transit cost | 0.012 | -0.258 | 0.163 | 0.091 | 0.234 | 0.910 |
| | (11.964) | (-168.122) | (43.078) | (30.214) | (89.670) | (69.969) |
| *Socio-demographic* | | | | | | |
| Age 30-45 | 0.019 | -0.006 | -0.083 | 0.151 | -0.062 | -0.166 |
| | (6.354) | (-1.366) | (-8.100) | (16.174) | (-8.655) | (-4.737) |
| Age 45-60 | 0.103 | -0.114 | -0.019 | 0.044 | -0.100 | -0.206 |
| | (29.682) | (-22.827) | (-1.567) | (4.120) | (-11.969) | (-5.578) |
| Age over 60 | -0.186 | -0.271 | -0.052 | 0.862 | 0.305 | 1.031 |
| | (-60.919) | (-8.557) | (-4.980) | (78.723) | (42.139) | (23.439) |
| Female | 0.024 | 0.022 | -0.011 | -0.017 | -0.081 | -0.143 |
| | (10.928) | (6.809) | (-1.459) | (-2.416) | (-15.455) | (-5.975) |
| Disb | -0.057 | -0.110 | -0.144 | 0.333 | 0.165 | 0.514 |
| | (-8.037) | (-10.764) | (-6.329) | (14.399) | (9.841) | (9.888) |
| Driving license | 0.055 | 0.010 | - | - | -0.086 | -0.044 |
| | (28.186) | (3.427) | | | (-18.089) | (-1.388) |
| Restricted car access | -0.082 | -0.057 | 0.119 | 0.035 | 0.032 | 0.186 |
| | (-32.677) | (-15.113) | (13.123) | (4.358) | (5.180) | (5.482) |
| Unrestricted car access | -0.097 | -0.061 | 0.266 | -0.078 | -0.058 | 0.191 |
| | (-30.612) | (-12.583) | (22.114) | (-7.795) | (-7.309) | (4.775) |
| Threshold1 (Bias1) | | | 6.466 | | | 1.360 |
| | | | | | | (26.030) |
| Threshold2 (Bias2) | | | -1.034 | | | 3.259 |
| | | | | | | (61.660) |
| Threshold3 (Bias3) | | | -4.663 | | | 4.967 |
| | | | | | | (83.049) |
| Threshold4 (Bias4) | | | -7.653 | | | 7.046 |
| | | | | | | (92.789) |
| No. observation | | | 45,547 | | | 45,547 |
| No. parameters | | | 422 | | | 17 |
| Log-likelihood | | | -14483.3 | | | -31169 |
| AIC | | | 29810.6 | | | 62372 |
| Validation accuracy | | | 81.89% | | | 55.97% |

*t-stats are outlined in the parenthesis

In regard to the thresholds and bias parameters, it is worth mentioning that the results of models cannot be interpreted in the same way. For instance, in the ordered logit model, if the utility function or latent variable is higher than threshold1 and less than threshold2, the pedestrian's waiting time is predicted to be medium. However, non-increasing bias parameters show that predicted probabilities are non-increasing, guaranteeing consistency between binary classifiers. In fact, the role of bias parameters is rooted in the deep learning formulation of our proposed model and there is no interpretable relationship between bias parameters and the utility function.

*5.2. The Effect of Explanatory Variables*

In addition to high prediction accuracy, interpretability distinguishes the ordinal-Reslogit model from other deep learning-based models for ordinal data. This important attribute provides the feasibility of analyzing the effect of observed variables known as controlling or targeting variables in travel behavioural models. In general, transportation and urban planners mainly focus on observed variables for making any changes in the transportation network. Therefore, in this section, we intend to demonstrate this attribute of Ordinal-ResLogit model by describing the impact of contributing factors on pedestrian wait time and travel distance. Since in recent years, the interaction between AVs and road users has been of paramount importance in transportation literature, in this section, we strive to analyze the effect of contributing variables in SP data in more detail than RP data, so as to highlight the probable changes in pedestrian behaviour in the presence of AVs.

*5.2.1. Pedestrian wait time (SP data)*

The impact of significant variables on three levels of pedestrians waiting time can be observed in Table 4 in which a positive coefficient means that the probability of waiting longer on the sidewalk is higher and for a negative coefficient, the reverse is true. Interestingly, highly similar results are obtained from the Ordinal-ResLogit model when compared to the study conducted by Kalatian and Farooq (2021) with the same data, who utilized a completely different deep learning structure. Based on the t-statistic information, rooted in the standard error of parameters, in the ordered logit model, the effects of some important variables such as high lane width, snowy weather, etc., are insignificant even in 90% level. However, these variables have highly significant effects on pedestrian waiting time in the Ordinal-ResLogit model. These issues underscore the further better performance of the Ordinal-ResLogit model in comparison to the ordered logit model.

The significant positive coefficient of mixed traffic conditions shows that pedestrians tend to wait longer in this condition to find a safe gap for crossing, compared to the current condition. By contrast, the fully automated situation significantly increases pedestrian preference towards low waiting times. As in the experiments, participants could distinguish human-driven from AVs and there is a general trust in braking systems of AVs, the negative effect of fully automated condition compared to fully human-driven condition is expected. It is notable that the ordered logit model is estimated a completely different effect for the fully automated variable.

Regarding road-related variables, wider lane width and higher density both highly significantly affect longer waiting times. As pedestrians constantly try to find an appropriate gap for crossing, high density decreases the opportunity for pedestrians to cross safely. Therefore, longer waiting times are likely to be selected. Moreover, intuitively, crossing from wider roads is more dangerous due to increasing the number of vehicles driven in different lanes, resulting in a higher level of waiting time. In addition, the results indicate that road type is a contributing factor in choosing waiting time alternatives. The level of waiting time decreases on the sidewalk of the two-way road with a median compared to a similar street without a median. However, crossing from a one-way road significantly increases the probability of choosing longer waiting times. Interestingly, in terms of lane width, similar results were obtained in previous studies (Rasouli et al., 2017).

Age groups have different effects on pedestrian waiting time such that in general, with increasing age, pedestrians are more cautious about crossing the street and wait longer on the sidewalk. Out of different age groups, the 18-29 group is settled as a baseline category. The significant positive coefficient related to pedestrians aged over 50 emphasizes that the eldest age groups prefer to wait longer on the sidewalk. Our result is in line with previous studies in the literature (Hamed, 2001; Sun et al., 2003). Regarding gender, the result demonstrates that females significantly would rather wait longer in comparison to men. In general, as females behave more cautiously, they tend to start crossing the street in a safe condition. Furthermore, having more than one car in the households can considerably decrease the probability of choosing low waiting time, meaning that the greater chance of using a private car leads to a greater perception of dangers, rooted in the interaction between vehicles and pedestrians.

Another noteworthy result of this study shows that the lifestyle influences the behaviour towards crossing the street. Pedestrians, who usually use active modes, wait longer at the sidewalk in comparison to those using a private car. However, a reverse result is obtained for people who usually use public transportation. The regular schedule of public transportation causes people to behave hurriedly while people using active mode are more patient for crossing the streets. In the real world, public transportation users tend to minimize the waiting time at transit stops and they try to reach the public modes on time. Similarly, if walking becomes a daily habit of pedestrians or people walk to work regularly, the probability of waiting short times decreases. It is worth noting that Kalatian and Farooq (2021) concluded the opposite results for the effect of walking to work variables. They traced back their results to higher comfort of pedestrians with regular walking habits. However, in the real world, people who walk daily, particularly in peak hours, are more aware of dangers and essential rules related to pedestrians. As a result, they are more likely to choose longer waiting times in comparison to others.

Our developed model significantly estimates the effect of environmental variables, snowy and night conditions. Snowy weather decreases the ability of pedestrians to detect distance, vehicle speed, etc., causing longer waiting times. Previous studies confirm that bad weather conditions adversely influence pedestrians' crossing behaviour (Sun et al., 2003). Nonetheless, the positive effect of the night variable on the low waiting time category is not similar to the snowy variable. Kalatian and Farooq (2021) mentioned that in their design of virtual experiments, the participants

'sight was not limited in scenarios with night conditions. In their virtual experiments, the mental effect of the night was simulated using changes in the color of the sky. Therefore, this opposite result between mentioned environment-related variables may root in the unreal condition in the night scenarios. Despite numerous advantages of the VIRE approach for collecting futuristic and dangerous scenarios, simulation of the real effect of lighting and weather condition is considered as a limitation of virtual reality experiments.

### 5.2.2. Travel distance (RP data)

According to the results of Ordinal-ResLogit for RP travel distance data, the effect of all travel-related and socio-demographic variables are highly significantly estimated, except the effect of driving license on medium and long categories. In the utility function of each ordinal category, a positive coefficient means that the probability of choosing that category is higher, and the reverse conclusion is true for a negative coefficient.

In comparison to private cars, using public transportation positively affects longer travel distances, although the positive impact of public transportation can be observed on choosing short travel distance. Conversely, people who prefer to regularly use active modes prefer to choose a closer destination. This result is in conjunction with physical limitation for walking or cycling. Another determining factor is travel cost such that higher travel cost encourages people to choose a farther destination for their non-mandatory trips. This result may be rooted in more flexibility in choosing shopping malls, restaurants, recreational places, etc.

Regarding the age groups, the results show that people aged over 60 more likely to choose longer distance; however, for other age groups, there is no consistent preference toward choosing travel distance. In other words, although the 30-45 and 45-60 age groups tend to have a short travel distance compared to 18-29, they also likely choose long distances for their trip. Moreover, our results show that females also prefer longer distances in comparison to men and surprisingly disabled people have more preference towards longer travel distance. In terms of the driving license, people who have this official document less prefer to travel to farther destinations. Similarly, people who readily have access to private cars in their household are likely to choose a shorter destination rather than a longer travel distance. These results may trace back to their regular transportation mode. In fact, as such group of people probably commute by private car, they prefer to choose shorter travel distance for their non-mandatory trips and use other transportation modes. While according to our results, people who hardly access a private car, likely choose a longer distance compared to baseline group who have no access to private car.

*5.3. Economic Information*

To evaluate the impact of various transport policies, analyzing the effect of a change in contributing factors on decision makers' choice has been another main goal of travel behavioural modelling. Since the emergence of combining deep neural networks with discrete choice models, the analysis of deep learning-based discrete choice models in terms of both explanatory variables and economic information has been under debate in the literature, albeit extraordinary achievement in prediction. In fact, similar to conventional behavioural models, researchers are interested in extracting reliable economic information from the novel learning-based travel behavioural models. This study aims to show that our proposed model can be as complete as the traditional ordered logit model. We try to derive various economic information from Ordinal-Reslogit model by exploiting the deterministic part of utility function. In other words, the main distinguishable characteristic of Ordinal-Reslogit model is utility function specification; therefore, not only does Ordinal-Reslogit model enjoy a deep learning process in prediction of individuals' ordinal preference, but also is as interpretable as the ordered logit model in terms of explanatory variables and economic information. Hence, the process of extracting economic information is completely similar to traditional DCMs and we can readily examine the impact of any changes in targeting factors on decision makers' choice. In the following section, three different types of economic information, market share, substitution pattern of alternatives and elasticity, are derived from the Ordinal-Reslogit model for both RP and SP data.

*5.3.1. Market share*

The market shares of ordinal alternatives are presented in Table 6 and 7 for both SP and RP data respectively. This economic information helps us to precisely evaluate the accuracy or performance of Ordinal-Reslogit in prediction. By comparing the market shares of different alternatives of SP data, we found that both Ordinal-ResLogit and ordered logit model suffer from lack of accuracy in prediction such that the market share of high waiting time in both models is zero. Conversely, similar to the accuracy of the estimated Ordinal-ResLogit model for RP data, the analysis of market shares also confirms that the market shares of ordinal travel distance alternatives predicted by our proposed model are very close to the true market shares. Nevertheless, the market shares of RP data obtained based on the results of ordered logit model emphasize that there is a considerable error between the predicted and actual percentage of market shares, resulting in 44.21% error in prediction. Therefore, the analysis of market shares of both datasets highlights that the less accuracy in SP data is not rooted in model specification of Ordinal-ResLogit, but data collection process. Interestingly, our proposed model performs significantly more accurately than the traditional ordered logit for RP ordinal data, which is more reliable than SP data.

Table 6: Market share of ordinal alternatives of SP data for both estimated models

| Ordinal Alternatives | Ordinal-ResLogit | Ordered Logit | Actual Market Share |
|---|---|---|---|
| *Pedestrian Wait Time (SP data)* | | | |
| Low | 77.6% | 81% | 64.4% |
| Medium | 22.4% | 19% | 29.4% |
| High | 0.0% | 0.0% | 6.3% |

Table 7: Market share of ordinal alternatives of RP data for both estimated models

| Ordinal Alternatives | Ordinal-ResLogit | Ordered Logit | Actual Market Share |
|---|---|---|---|
| *Travel Distance (RP data)* | | | |
| Very short | 47.3% | 56.5% | 46.6% |
| Short | 27.4% | 28.8% | 28.1% |
| Medium | 16.8% | 8.3% | 14.9% |
| Long | 6.9% | 4.7% | 7.4% |
| Very long | 1.6% | 1.6% | 3.0% |

*5.3.2. Substitution pattern of alternatives*

The substitution pattern of the alternatives is another important choice analysis that provides deep insight into the impact of a targeting variable on the choice probability of all alternatives, while holding all other contributing variables constant. In other words, this economic information assists us in the prediction of policies effects on market shares. Regarding pedestrians wait time, we seek to understand how the choice probability of alternatives varies as the road density increases and for travel distance data, the choice probabilities of ordinal alternatives are examined by varying driving cost and public transport cost. Figure 3, and 4 visualize the effect of these variables on the choice probability of alternatives.

Concerning pedestrian wait time (Figure 3), the alternatives in a choice set are not substituted, meaning that by increasing road density, individuals still choose a low waiting time. A similar result is obtained for the traditional ordered logit model with a slight difference. Although according to the Ordinal-ResLogit model, by increasing road density, the choice probability of medium wait time is still lower than low category, in the ordered logit model, when density is larger than 38 vehicles per kilometer, the medium level substitutes low waiting time. In other words, based on our proposed model, increasing road density does not affect pedestrian behavior before starting crossing.

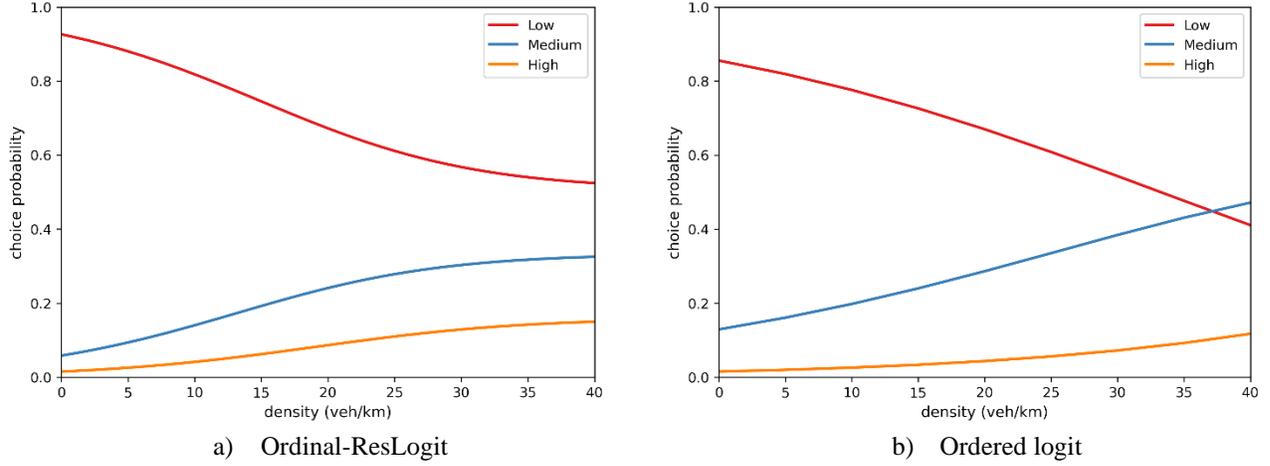

a) Ordinal-ResLogit　　　　　　　　　　b) Ordered logit

Figure 3: Substitution patterns of alternatives of SP data (pedestrian wait time) with varying density

As shown in Figure 4a, in the ordered logit model, more driving cost leads to longer travel distance such that higher than cost of $23, very long alternative substitute for other categories. However, we can observe more complex behavior based on Ordinal-Reslogit. More precisely, when the driving cost is between $2 to $5, there is more preference towards choosing long travel distance, but between $5 and $12, the medium category again substitutes for long travel distance. However, when the driving cost becomes higher than $12, the substitution pattern illustrates that the medium and short alternatives are substitutes to each other. Therefore, in the Ordinal-Reslogit model, there is no positive constant impact between driving cost and travel distance and the effect of any changes in driving cost considerably depends on the amount of changes. A similar result can be observed for transit costs. While substitution pattern is less sensitive to changes since the result shows that approximately after 20$ and $28, short and very long distances substitute for other categories, respectively. Generally, in choice analysis, individuals are expected to change their decision on their destination of non-mandatory trips, as the travel cost increases and the Ordinal-ResLogit model better reflect this effect.

### 5.3.3. *Elasticity*

Elasticity is the most informative choice analysis for understanding the impact of a change in contributing factors on decision makers' choice. Mathematically, elasticity is the standard form of partial choice probability derivatives of each individual with respect to a controlling or targeting variable calculated as follows:

$$E_{x_{in}}^{j} = \frac{\partial P_n^{\,j}}{\partial x_i} \times \frac{x_{in}}{P_n^{\,j}} \tag{15}$$

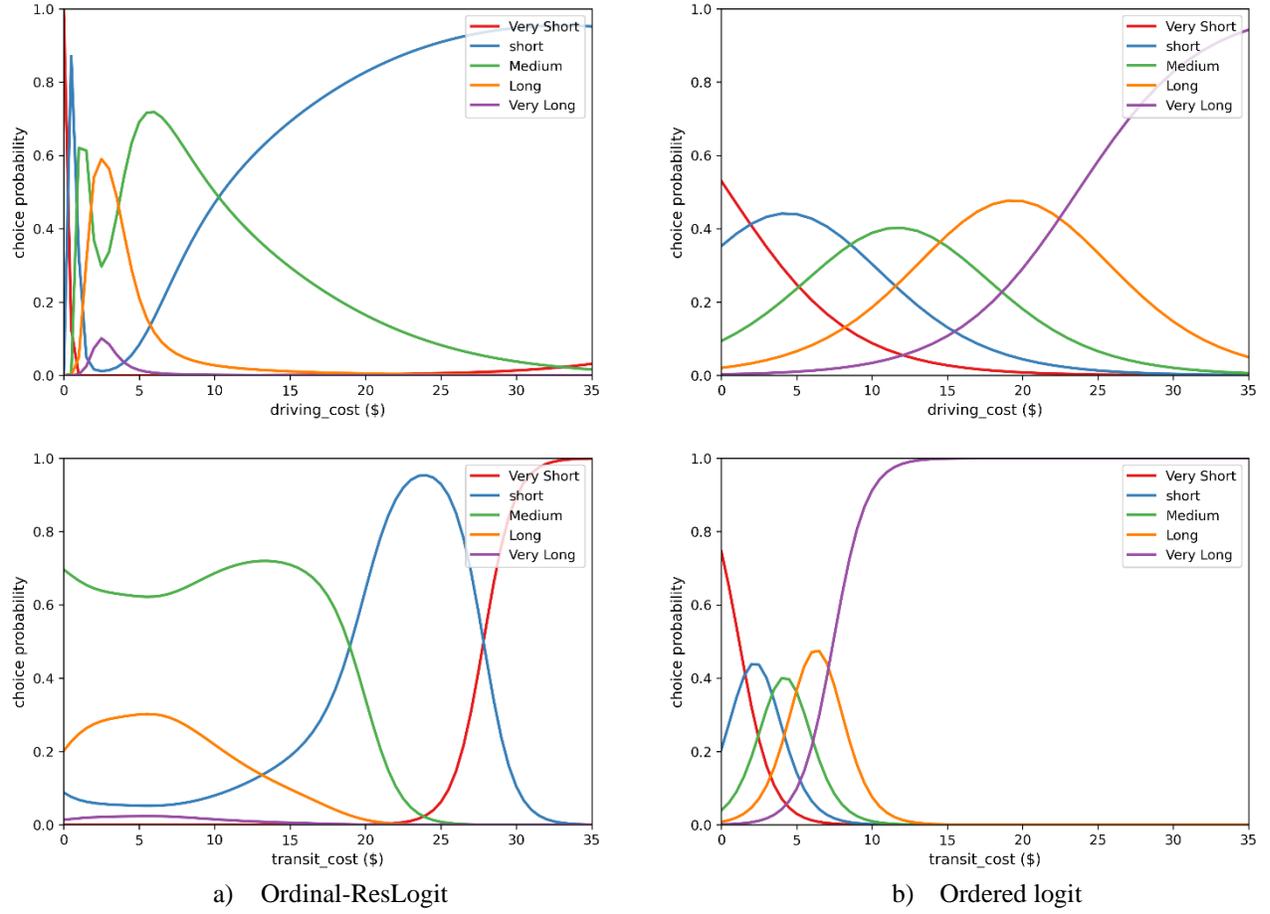

a) Ordinal-ResLogit          b) Ordered logit

Figure 4: Substitution patterns of alternatives of RP data (travel distance) with varying travel cost

$E_{x_{in}}^{j}$ is disaggregate form of elasticity of alternative $j$ with respect to variable $x_i$ and the aggregate elasticity, $E_{x_i}^{j}$, is (Ben-Akiva, M.E., Lerman, 1985):

$$E_{x_i}^{j} = \frac{\sum_{n=1}^{N} P_n^{\ j} \times E_{x_{in}}^{j}}{\sum_{n=1}^{N} P_n^{\ j}} \tag{16}$$

Where $P_n^{\ j}$ is the probability of individual $n$ choosing alternative $j$ from the choice set. Elasticity is usually used for continuous variables that 1% increase or decrease in the variable is meaningful. However, as in this study, the most contributing factors are not continuous, we also employ another method for evaluating elasticity of binary variables. In general, for binary variables like gender, there are two possible values {0: Female, 1: Male}. To obtain the effect of changes in the value of binary variables on individuals' behaviours, First, the value of the variable for each individual is converted to 0 or 1 based on the initial status. Then, by considering these changes, the variation of choice probability is calculated for each individual. For instance, if an individual $n$ is Female, by converting the value of gender variable from 0 to 1, her choice probability changes will be obtained (Sener & Bhat, 2011).

Moreover, for discrete forms of pedestrian wait time and travel distance data which inherently there are an ordinal relationship among alternatives, we are interested in understanding how changes in binary variables affect expected value of responses. To obtain the expected value of dependent variables, first, an integer is contemplated to each level of ordinal responses in such a way that each integer represents each level of ordinal responses. In this study, based on the result of discretization resulting from Jenks Natural Breaks method and logical facts, the average value of thresholds of different categories is assumed as the representative of each level of ordinal alternatives. Then, the expected value of a dependent variable for each individual can be obtained using equation (17):

$$E(Y_n) = \sum_{j=1}^{J} C_j \times P_n^{\,j} \tag{17}$$

Where $C_j$ is the representative of level of alternatives. In fact, the variation of expected value demonstrates how plans or actions affect the expected value of responses.

Table 8 and 9 present the elasticities of pedestrian wait time and travel distance with respect to targeting variables respectively. Each value represents the average elasticity of the respondents based on the two estimated models. Regarding SP data, the aggregate elasticity of road density is reasonable since 1% increase in density leads to a rise in pedestrian waiting time; however, this effect for the high category is considerably higher than the medium level. It is of note that the results of our proposed model and the ordered logit model are approximately similar in terms of both sign and magnitude.

Table 8: Elasticities of ordinal alternatives of SP data with respect to targeting variables

|  | **Low** | **Medium** | **High** |
|---|---|---|---|
| **Ordinal-ResLogit model** | | | |
| Density | -0.307 | 0.560 | 0.868 |
| **Ordered logit model** | | | |
| Density | -0.360 | 0.548 | 1.135 |

Concerning RP data, the ordered logit model indicates that 1% change in transit cost likely leads to the higher probability of choosing a farther destination for non-mandatory trips. However, in the Ordinal-ResLogit model, the magnitude of this elasticity is considerably smaller than the ordered logit model. The reverse result is obtained for the driving cost in terms of magnitude elasticity. Furthermore, the elasticity result of the Ordinal-Reslogit model shows that not having a driving license causes longer travel distances for non-mandatory trips, but with a considerably slight impact. In other words, in our choice analysis, if all individuals would not have a driving license, on average, the probability of choosing longer distances is increased. As shown in Table 9, in our proposed model the magnitude of this elasticity is smaller than the traditional ordered logit model. By considering the average thresholds of each category and computing the expected value, we conclude that not having a driving license averagely increases travel distance by 6.9% and 22.1% in Ordinal-ResLogit and ordered logit models respectively.

Table 9: Elasticities of ordinal alternatives of RP data with respect to targeting variables

|  | **Very short** | **Short** | **Medium** | **Long** | **Very long** |
| --- | --- | --- | --- | --- | --- |
| **Ordinal-ResLogit model** | | | | | |
| Driving cost | -0.821 | 0.512 | 0.948 | 1.160 | 1.723 |
| Transit cost | -0.023 | -0.008 | -0.101 | 0.130 | 0.782 |
| Driving license | 0.012 | 0.020 | -0.014 | -0.003 | 0.016 |
| **Ordered logit model** | | | | | |
| Driving cost | -0.081 | -0.014 | 0.014 | 0.212 | 0.923 |
| Transit cost | -0.374 | 0.006 | 0.453 | 0.875 | 1.822 |
| Driving license | -0.753 | 0.258 | 0.320 | 0.293 | 0.601 |

## 6. Conclusion

We propose a novel deep learning-based ordered logit model called an Ordinal-ResLogit model. The main purpose of our study is to develop an interpretable machine learning-based discrete choice model for ordinal data as well as achieving rank-monotonicity among ordinal responses. The formulation of the Ordinal-ResLogit model is derived from ResLogit model using Residual Neural Networks or ResNets architecture and CORAL framework, which is an ordinal learning algorithm. Our methodological contribution is a new ordered model proposed based on both traditional discrete choice models and deep learning algorithms.

The Ordinal-ResLogit model is able to overcome a general restriction of machine learning algorithms and deep neural networks known as black-box. The Ordinal-ResLogit model benefits from learning residual network or the ResNets algorithm such that this structure can capture unobserved behavioural heterogeneity, while retaining the deterministic component of the utility function, providing the possibility of econometric analysis. Therefore, our proposed model simultaneously enjoys learning process and interpretability and this attribute distinguishes Ordinal-ResLogit model from other deep learning frameworks used for ordinal data. Regarding ordinal responses, another important attribute of our proposed model is that although the Ordinal-ResLogit model is classified as a binary decomposition approach for ordinal data, this model does not suffer from inconsistency among classifiers. Moreover, the Ordinal-ResLogit framework does not require a misclassification cost matrix. In fact, in contrast to other classification problems, the cost values are not the same for different prediction error in ordinal classification. On the other hand, the formulation of the cost matrix strongly influences the performance of the model. Hence, being free from the cost matrix is another main attribute of our proposed model.

The parameters of Ordinal-ResLogit model are estimated through a data-driven approach by which we are able to estimate a large number of parameters of machine learning models. A mini-batch stochastic gradient descent algorithm is used to minimize the error of the validation set. In addition, due to skipped connection structure of ResNets architecture, each residual layer of our neural network is estimated independently, leading to an identifiable model structure. In other words, our proposed model does not suffer from the vanishing gradient problem during estimation.

To evaluate the performance of the Ordinal-ResLogit model, we use two types of dataset, 1) an advanced Stated Preference (SP) data collected by a virtual reality simulation framework and 2) a Revealed Preference (RP) data. With SP data, the ordinal categories of pedestrian wait time in the presence of Automated Vehicles (AVs) on the sidewalk is investigated. Although the majority of studies in this area are conducted by approaches in which continuous pedestrians' waiting time is analyzed, we concentrated on an ordinal discrete form of waiting time, comprising low, medium, and high waiting time. Therefore, another contribution of this study is to assess effect of mixed traffic conditions with the upcoming AVs on the road and fill the void of investigating pedestrian wait time employing deep learning ordinal models. However, generally, due to lack of complete reliability towards designing hypothetical scenarios, we also evaluate the structure of Ordinal-ResLogit by categorizing a travel demand RP data and using ordinal discrete form of travel distance. In fact, real travel behaviour information provides more reliable conclusions about the performance of our proposed model.

The results demonstrate that the Ordinal-ResLogit model outperforms the conventional ordered logit model for both RP and SP data. Particularly, in RP data, there is a considerable difference between the accuracy of the two estimated models, approximately 26%. Therefore, the Ordinal-Reslogit allows modellers to achieve more accurate predictions by capturing heterogeneity and the impact of unobserved variables. In terms of explanatory variables our proposed model is able to consider the effect of more variables for the prediction of pedestrian waiting time, while similar performance is not obtained from the conventional ordered logit model. Results of the Ordinal-ResLogit model further confirm that the presence of automated vehicles in future roads has a significant effect on pedestrian waiting time. More precisely, mixed traffic conditions increase the pedestrian propensity to choose longer waiting times. Conversely, results obtained for fully automated conditions are entirely different. Moreover, in this study, we aim to show that Ordinal-ResLogit model can be as complete as the traditional ordered logit model. We try to derive various economic information from Ordinal-Reslogit model by exploiting the deterministic part of utility function.

As we do not consider the regularization methods, enhancing the performance of current Ordinal-Reslogit by taking this step into account can be one future direction. In terms of evaluation, as there is a growing interest in VR-based SP data, the performance of our model can be analyzed based on a more extensive and realistic VR dataset. For instance, in our study, the impact of night conditions was not logical due to the design of the VR data. Furthermore, as there is a growing trend in modelling towards the combination of deep learning algorithms with existing choice models, further researchers can make a comparison between the performance of the Ordinal-ResLogit model and other learning-based ordinal models.

**Acknowledgments**

The financial support of the Natural Sciences and Engineering Research Council of Canada (NSERC grant #RGPIN-2020-04492) and Canada Research Program (grant # CRC-2017-00038) is gratefully acknowledged.